# Combining Subgoal Graphs with Reinforcement Learning to Build a Rational Pathfinder

**Junjie Zeng, Long Qin \*, Yue Hu, Cong Hu and Quanjun Yin**

College of System Engineering, National University of Defense Technology
zengjunjie13@nudt.edu.cn (J.Z.); huyue.nudt@gmail.com (Y.H.); hccz95@163.com (C.H.);
yin_quanjun@163.com (Q.Y.)
\* Correspondence: qldbx2007@sina.com;

**Abstract:** In this paper, we present a hierarchical path planning framework called SG–RL (subgoal graphs–reinforcement learning), to plan rational paths for agents maneuvering in continuous and uncertain environments. By "rational", we mean (1) efficient path planning to eliminate first-move lags; (2) collision-free and smooth for agents with kinematic constraints satisfied. SG–RL works in a two-level manner. At the first level, SG–RL uses a geometric path-planning method, i.e., Simple Subgoal Graphs (SSG), to efficiently find optimal abstract paths, also called subgoal sequences. At the second level, SG–RL uses an RL method, i.e., Least-Squares Policy Iteration (LSPI), to learn near-optimal motion-planning policies which can generate kinematically feasible and collision-free trajectories between adjacent subgoals. The first advantage of the proposed method is that SSG can solve the limitations of sparse reward and local minima trap for RL agents; thus, LSPI can be used to generate paths in complex environments. The second advantage is that, when the environment changes slightly (i.e., unexpected obstacles appearing), SG–RL does not need to reconstruct subgoal graphs and replan subgoal sequences using SSG, since LSPI can deal with uncertainties by exploiting its generalization ability to handle changes in environments. Simulation experiments in representative scenarios demonstrate that, compared with existing methods, SG–RL can work well on large-scale maps with relatively low action-switching frequencies and shorter path lengths, and SG–RL can deal with small changes in environments. We further demonstrate that the design of reward functions and the types of training environments are important factors for learning feasible policies.

## 1. Introduction

In this paper, we focus on the problem of planning rational paths in continuous and uncertain environments. By "rational", we mean that, firstly, computational costs, mainly the time consumed, brought by the planning algorithm must be low enough, so requirements such as avoiding first-move lags and providing human users with an excellent experience can be met; secondly, resultant paths must be collision-free and smooth, and thus, feasible to follow given kinematic constraints. One typical application of this problem lies with agents who maneuver in time-sensitive scenarios (e.g., service robots working in the real world and Non-Player Characters (NPCs) engaging movement tasks in video games). Briefly, application fields such as robotics and video games strongly require time-saving path-planning methods which can quickly generate realistic paths.

The problem of path planning is extensively researched in the literature [1–6]. A* on grid maps [6] is regarded as a fundamental and standard path-planning algorithm based on search. It is complete and optimal, which means A* can find an optimal path if there is at least one from the start point to the goal point. There are at least two shortcomings of A* on grid maps: (1) path symmetry [3], i.e., the existence of symmetrical paths forces A* to generate same states multiple times along different paths, which consumes more time; (2) cell-wise expansion, i.e., since successors of a cell which are expanded are just its adjacent cells, this short-ranged expansion brings about a huge open list and a large number of operations. These two problems cause many unnecessary cell generations and expansions, and, as a result, A* is so slow that it cannot meet the first requirement for rational paths.

To overcome the above disadvantages, methods for map abstraction attract many researchers' attention. Subgoal graphs (SG) [5] is an archetypical map-abstraction-based algorithm, whose preprocessing stage is responsible for building subgoal graphs. Similar to visibility graphs, its simple version—SSG—places subgoals at the corners of obstacles. SSG links cells, between which a reachability relationship called direct-h-reachability is satisfied, and all symmetrical paths are traversable. Then, over subgoal graphs, A* is used to obtain the abstract path. Finally, depth-first search is safely utilized to refine it into a ground-level path. As a result, A* on subgoal graphs works far better than that on grids and satisfies the first requirement mentioned above. As for the second requirement, however, the optimal ground-level path is constrained to grid edges (i.e., the heading changes are artificially constrained to specific angles) [7], which causes sharp turns in the generated paths, and cannot meet kinematic constrains. Moreover, it is time-consuming for SG to deal with changing environments by reconstructing subgoal graphs.

To plan kinematically feasible paths, interpolating curve planners [8] construct and insert a new set of data within the range of a previously known set. For example, in Reference [9], hypocycloidal curves were used to smooth sharp turns, which can maintain a safe clearance in relation to the obstacles. However, when unexpected obstacles often appear in the detection range of robots, these kinds of planners need to recalculate a new path frequently, causing more time consumption. It is also difficult for sampling-based planners, such as Probabilistic Roadmaps (PRM) [10] and Rapidly exploring Random Trees (RRT) [11], to deal with unexpected obstacles. Furthermore, the generated paths are random and suboptimal.

To deal with uncertain environments and improve the path quality, Reinforcement Learning (RL) can be used to learn near-optimal policies to find paths in the real world under certain constrains. RL is learning what to do—how to map situations to actions—to maximize a numerical reward signal [12]. Specifically, RL is an agent that interacts with the environment, and learns an optimal policy by trial and error [13]. RL emerged as a practical method for robot control [14] and was successfully applied to solve complex robot manipulation problems [15] and learn complex skills like playing Go [16] and Atari games [17]. There are two categories of RL (i.e., model-based RL and model-free RL). Since we cannot provide complete models of uncertain environments for RL, model-free RL is more suitable for our scenarios. However, RL cannot be used directly to plan paths in complex environments due to two limitations: (1) sparse reward, i.e., reward is sparsely distributed in large state spaces, causing difficulties for learning feasible policies; (2) local minima trap, i.e., RL agent may be trapped in local minima, like goals on the other side of box canyons.

In this paper, to overcome the abovementioned limitations of SG and RL methods, we present a hierarchical path planning framework, called SG–RL, to integrate the geometric path-planning method (i.e., SG) and the ground-level motion planning method (i.e., RL), which can plan rational paths in uncertain and continuous environments. SG–RL solves this path-planning problem in a two-phase manner. The first phase is a global abstract path-planning phase that focuses on the first requirement for rational paths. The second phase is a feasible trajectory-planning phase that focuses on the second requirement. In the first phase, SG–RL uses SSG to find global abstract paths, also called subgoal sequences, from start points to goal points with high computational efficiency. In the second phase, considering kinematic constraints, SG–RL uses an RL method, i.e., Least-Squares Policy Iteration (LSPI) [18], to learn near-optimal motion-planning policies which can generate kinematically feasible and collision-free trajectories between adjacent subgoals.

Compared with RL methods, SG–RL can use SSG to plan paths more efficiently so as to eliminate first-move lags, and owing to the direct-h-reachability of adjacent subgoals, SG–RL can support RL agents to realize long-range navigation on complex maps by overcoming sparse reward and local minima trap. Then, compared with SG methods, SG–RL can use LSPI to generate smooth paths that follow kinematic equations and preserve G1 continuity [9], and due to the generalizability of LSPI, SG–RL can deal with small changes (i.e., unexpected obstacles) over maps.

The rest of this paper is organized as follows: Section 2 discusses some related work on path-planning methods based on A*, RL, and hierarchical path planning. Section 3 presents the hierarchical path-planning approach based on SG and RL. In Section 4, the performance of the proposed approach is evaluated by simulation experiments. Finally, this paper is concluded in Section 5.

**2. Related Work**

In this section, studies of three kinds of methods (i.e., path-planning methods based on A*, RL, and hierarchical path planning) in the path-planning field are introduced.

*2.1. Path-Planning Methods Based on A\**

Hart, Nilsson, and Raphael jointly proposed the A* algorithm, which is a widely used best-first search algorithm and can search an optimal path over grid maps [6]. A* plays an important role in areas that do not require real-time response. However, with many real-time applications getting prevalent in fields such as real-time strategy games and robotics, A* faces severe challenges (e.g., searching in large scale and dynamic environments). A* tends to scale poorly as it must compute complete and optimal paths for agents before agents can move, which brings the first-move lag problem. In brief, A* on grid maps has two mentioned weaknesses (i.e., path symmetry and cell-wise expansion). These two weaknesses cause a huge search space and unnecessary cell generations, which renders A* very slow.

To solve these problems, some recent studies focusing on map representations were proposed. Such kinds of studies mainly construct particular map representations via abstracting topological structures and key information of original maps to reduce the search space. Jump point search (JPS) [2] proposed by Harabor and Grastien uses the canonical ordering method to solve the path symmetry problem and identifies jump points over grid maps to reduce the search space. Then, the optimal canonical path is searched among jump points. However, in JPS, jump points do not really "jump" to one another; instead, they just "roll" between jump points, namely to check cells one by one online. To increase the search speed between jump points, JPS+ [3] builds jump-point maps via preprocessing. Since the distance information regarding adjacent jump points exists in the jump-point map, the path between jump points can be directly found. There is another kind of method named Contraction Hierarchy (CH) [4] that can search over grid maps faster than JPS+ [19], due to its distinctive preprocessing. The preprocessing of CH involves the contraction of one node at a time out of the graph and adds shortcut edges to the remaining graph. However, CH costs more preprocessing time and memory space than JPS+. SSG [5], proposed by Tansel Uras and Sven Koenig, constructs subgoal graphs by identifying subgoals and checking direct-h-reachability in the preprocessing stage. Compared with the original map, the subgoal graph abstracts the information of key points, so as to reduce the search space. The direct-h-reachability ensures that any shortest paths between adjacent subgoals are not blocked by obstacles, allowing the path symmetry problem can be solved. To reduce the size of subgoal graphs, the two-level subgoal graphs method [5] relaxes connection conditions between subgoals and adds a subgoal-pruning strategy. To further increase the search speed, the n-level subgoal graphs approach [20] was proposed. These above path-planning methods based on A* can meet the first requirement of rational paths. However, these methods cannot solve the ground-level motion-planning problem under certain constraints and cannot deal with changes of environments.

*2.2. Reinforcement Learning*

In the field of ground-level motion planning, RL recently gained prevalence for systems with unknown dynamics [14]. Q-learning and Sarsa [12] can be useful for dealing with discrete state spaces. For example, Mihai Duguleana et al. [21] combined Q-learning with the artificial neural network for solving the problem of autonomous movement of robots in environments that contain both static and dynamic obstacles. However, in large or continuous state spaces, the abovementioned tabular RL methods are inefficient or impractical for applications. Function approximation methods that estimate value function can be used to tackle this problem [22]. There are many function approximators, such as artificial neural networks, decision trees, linear basis functions, and so on. In this paper, we focus on linear-basis-function approximators. The Least-Squares Temporal-Difference learning (LSTD) algorithm [23] proposed by Bradtke and Barto, an ideal method for prediction problems, uses linear basis functions to approximate the value function of a fixed policy. Though LSTD has many good properties like data efficiency and fast convergence, it cannot be straightforwardly used to learn good control policies. To solve this problem, the LSPI [18] algorithm proposed by Lagoudakis and Ronald introduces Least-Squares Temporal-Difference learning for the state-action value function (LSTD$Q$) method. LSTD$Q$ is an algorithm similar to LSTD and can learn the least-squares fixed-point approximation of the state-action value function of a policy via linear basis functions. LSPI adds LSTD$Q$ into the approximate policy-iteration structure, allowing it to get an optimal policy quickly by combining the policy-search efficiency of policy iteration with the data efficiency of LSTD$Q$. Due to its generalizability, LSPI can handle changes in environments. In this paper, we choose LSPI as the RL method. However, RL cannot be used directly in path planning in complex environments due to two limitations: (1) sparse reward; and (2) local minima trap.

*2.3. Hierarchical Path Planning*

To overcome limitations of path-planning methods based on A* and ground-level motion-planning methods, hierarchical path-planning approaches are widely researched in the literature [24–28]. In Reference [24], a two-level-based, goal-driven architecture is used to solve mobile robot navigation in real life with vision systems and infrared radiation sensors. In Reference [25], a two-stage path-planning algorithm uses a variant of A* search to obtain a kinematically feasible trajectory in the first stage and improves the quality of the solution via numeric non-linear optimization in the second stage, for an autonomous vehicle operating in an unknown semi-structured (or unstructured) environment where obstacles are detected online by a robot's sensors. In Reference [26], to guarantee the consistency between high and low levels of planning, Cowlagi and Tsiotras proposed a hierarchical motion-planning framework with a novel mode of interaction between the geometric path planner and the vehicle trajectory planner. In Reference [27], a novel hierarchical global path-planning approach for mobile robots in a cluttered environment with multi-objectives was proposed. This approach has a three-level structure which combines triangular decomposition, Dijkstra's algorithm, and a proposed particle swarm optimization called constrained multi-objective particle swarm optimization. In Reference [28], to find an optimal maneuver that moves a car-like vehicle between two configurations in minimum time, a two-phase algorithm firstly solves a geometric optimization problem and then finds the optimal maneuver with system dynamics and its constraints satisfied.

In this paper, to overcome the abovementioned disadvantages of SSG and LSPI, we designed a suitable hierarchical framework to combine a global abstract path planner based on SSG with a feasible trajectory planner based on LSPI, which can improve the computational efficiency of hierarchical path-planning methods by optimizing the relationship between high and low levels of planning.

**3. Materials and Methods**

In this section, the problem description and the tracked robot used in the simulation experiments are briefly introduced in the first subsection. Next, the architecture of the proposed hierarchical path-planning method—SG–RL—is presented. Then, the SG-based global abstract path planner and the LSPI-based feasible trajectory planner are described in detail.

*3.1. Problem Description*

The path-planning problem investigated in this paper can be stated as follows: in continuous and uncertain environments, the problem is to plan rational paths for tracked robots from a start position to a goal position.

In this paper, we used the Cartesian coordinate system. The schematic diagram of the tracked robot is shown in Figure 1. The position state of the tracked robot in this paper is defined as $s = (x_r, y_r, \theta)$ by the global coordinate related to the map, where $x_r$ and $y_r$ are the tracked robot's abscissa and vertical coordinates, respectively, and $\theta$ is the angle between the forward orientation of the tracked robot and abscissa axis. We equipped six ultrasonic sensors at the front of the tracked robot to perceive the external environment. The detection angle of each ultrasonic sensor was 30 degrees. The max detection distance of each sensor was $D_{max}$. The tracked robot was driven by the left and right tracks. The kinematic equations of the tracked robot were

$$\begin{pmatrix} \dot{x} \\ \dot{y} \\ \dot{\theta} \end{pmatrix} = R \begin{bmatrix} \frac{1}{2\cos\theta} & \frac{1}{2\cos\theta} \\ \frac{1}{2\sin\theta} & \frac{1}{2\sin\theta} \\ \frac{1}{l_B} & \frac{-1}{l_B} \end{bmatrix} \begin{pmatrix} \omega_l \\ \omega_r \end{pmatrix}, \qquad (1)$$

where $R$ is the radius of the driving wheels in the tracks, $\omega_l$ and $\omega_r$ are the angular velocity of left tracks and right tracks (rad· $s^{-1}$), respectively, and $l_B$ is the distance between left tracks and right tracks.

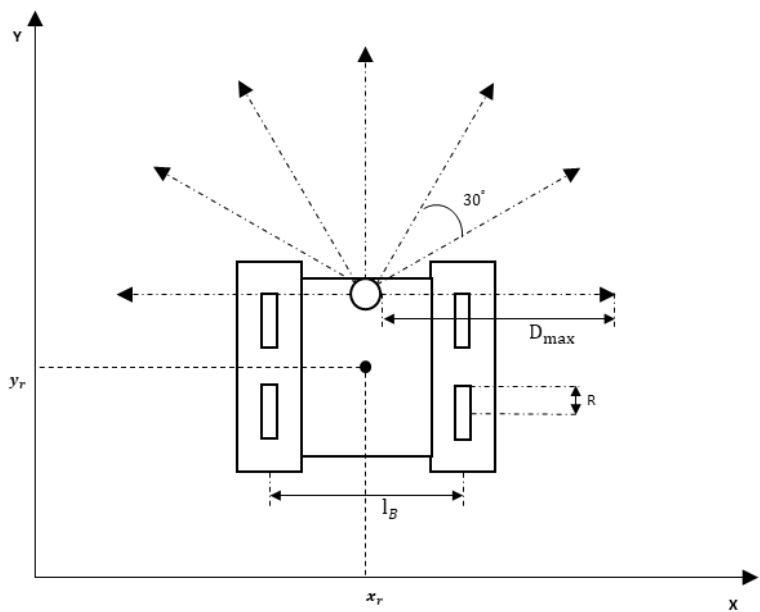

**Figure 1.** Schematic diagram of the tracked robot.

*3.2. The Architecture of the Proposed Algorithm*

As mentioned above, we decomposed the complex path-planning problems into two phases: the first for the global abstract path planning and the second for the feasible trajectory planning. In the first phase, we used SSG to find global abstract paths with low computation costs. In the second phase, we chose LSPI as the RL method. Then, a feasible trajectory planner based on LSPI took subgoals, which are cells of abstract paths, as input, and planned collision-free trajectories between subgoals with kinematic constraints satisfied. Moreover, there were online stages and offline stages in both phases. Figure 2 shows an overview of this approach.

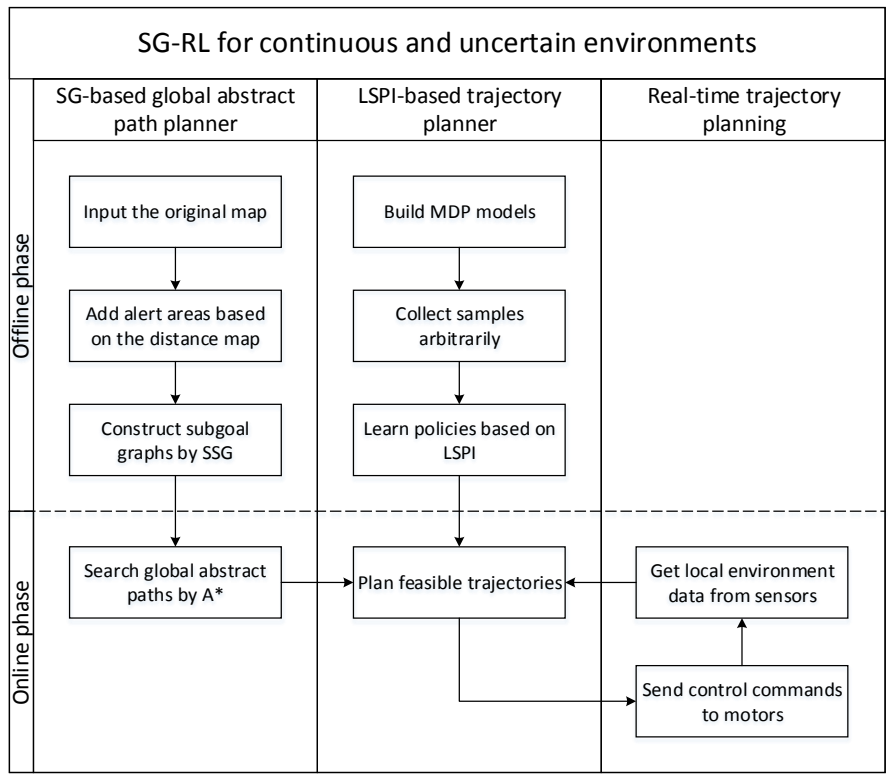

**Figure 2.** Subgoal graphs–reinforcement learning (SG–RL) flowchart. SSG: simple subgoal graphs; LSPI: least-squares policy iteration.

The main task of the first phase was providing subgoal sequences which began with the start point and ended with the goal point. During the offline stage, firstly, considering sizes of robots and user-defined safety distance, we added alert areas into original maps based on the distance map to plan safe and feasible paths. Then, we used SSG to construct subgoal graphs from modified maps by placing subgoals at the corners of obstacles and adding edges between related subgoals. During the online stage, we could use A* to search global abstract paths over the subgoal graph. It was not necessary to obtain the optimal ground-level paths between subgoals and assemble them into a complete one. Since basic search methods can hardly generate smooth and feasible solutions, SG–RL passes the task to the RL phase.

The main task of the second phase was planning feasible trajectories between subgoals. During the offline stage, firstly, this task can be divided into two sub-tasks which involve approaching subgoals with kinematic constraints and avoiding initially known or unknown obstacles. By analyzing the key information abstracted from sub-tasks, two Markov Decision Processes (MDPs), which were built separately, formally describe an environment for the RL method, LSPI. Next, according to MDPs, we collected samples arbitrarily from training environments which were smaller than test environments. Then, by carefully tuning parameters, LSPI could train feasible trajectory planners for robots. During the online stage, taking subgoal sequences and local sensor data as input, trajectory planners could choose the best action to generate collision-free and kinematically feasible paths. Note that LSPI uses the original map as the working environment, since alert areas were built by us, and the robot cannot recognize alert areas from local sensor data.

The first strength of SG–RL is that the excellent performance of SSG can be fully utilized, and some limitations of applying LSPI to plan paths in complex environments can be solved by SSG. (1) SSG uses A* to efficiently plan subgoal sequences over subgoal graphs constructed based on maps with alert areas, which can eliminate first-move lags. (2) SG–RL does not select subgoals from the optimal geometrical path according to certain standards [24]; however, it utilizes the intermediate, the abstract path regarded as subgoal sequences, which greatly increases the speed of providing subgoal sequences. Subgoals are placed at the exit of the local minima area, and the adjacent subgoals are direct-h-reachable, which ensures that all symmetrical paths between them are traversable, so as to eliminate the local minima trap for LSPI. (3) SG–RL uses SSG to generate subgoal sequences that are the input of LSPI. Since the distance between adjacent subgoals is shorter than the distance between the start and goal point, the reward distribution becomes relatively dense, which solves the limitation of sparse reward for LSPI. In conclusion, SSG used in SG–RL not only meets the first requirement of rational paths, but also overcomes the two abovementioned limitations of LSPI.

The second strength is that, based on optimal abstract paths over discretized grid maps, LSPI can train near-optimal feasible trajectory-planning policies which can plan collision-free and smooth continuous paths, following kinematic equations. Owing to linear-basis-function approximators, policies can take continuous state spaces as input to deal with continuous environments. Consequently, SG–RL satisfies the second demand for rational paths.

The third strength is that the SG–RL method can quickly adapt to uncertain environments via the generalizability of LSPI. When the map changes slightly, SG needs to reconstruct all subgoal graphs and replan paths. The constructed subgoal graphs cannot be directly applied to changing environments. Until now, there is no research regarding locally reconstructing subgoal graphs. Therefore, SG–RL uses the generalizability of LSPI to deal with small changes in environments based on original subgoal sequences without reconstructing subgoal graphs.

In conclusion, SG–RL not only makes good use of outstanding advantages of SG and RL, but also overcomes each limitation, allowing it to plan rational paths in continuous and uncertain environments.

*3.3. The SG-Based Global Abstract Path Planner*

Figure 3 shows the offline work concerning building alert areas and constructing subgoals. Since paths found by SSG over the original map are close to obstacles, they are not safe for robots to follow. To tackle this problem, we added alert areas into original-map-based distance maps.

During the phase of building alert areas, the dynamic brushfire algorithm [29] was used to build the Euclidean distance map (Figure 3b) based on the original map (Figure 3a). Then, the map with alert areas was constructed based on the Euclidean distance map. The size of alert areas was determined by the sizes of the robot and user-defined safety distance. One distinctive advantage of using a distance map to build alert areas is that one distance map can be used to construct various sizes of alert areas.

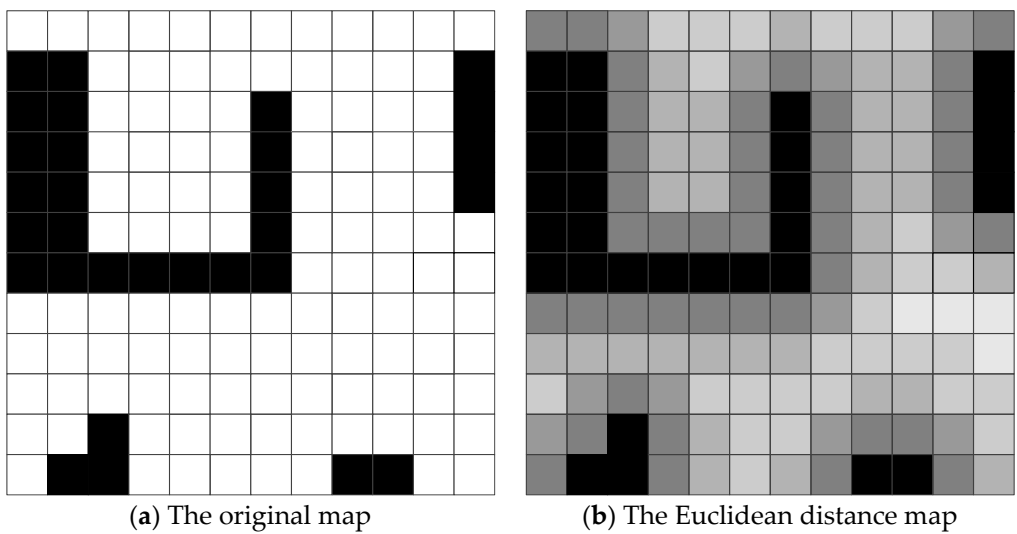

(**a**) The original map    (**b**) The Euclidean distance map

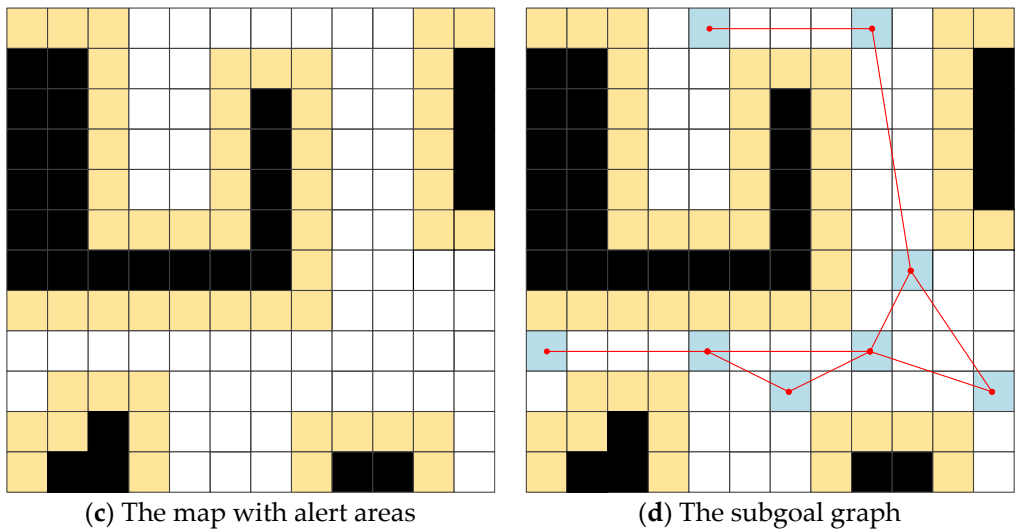

(**c**) The map with alert areas  (**d**) The subgoal graph

**Figure 3.** (**a**) This is the original map. (**b**) Black represents occupied cells. The brightness of cells increases with distance. (**c**) The light-yellow cells denote the alert areas. (**d**) The light-blue cells represent subgoals, and the red lines signify the direct-h-reachability of adjacent subgoals.

Then, we used the SG algorithm to find abstract paths on maps with alert areas and provide subgoal sequences for the feasible trajectory planner. There are three kinds of SG algorithm, namely simple, two-level, and n-level subgoal graphs [5,20]. We chose SSG as the method at this phase because, in SSG, all symmetrical paths between adjacent subgoals are not blocked by obstacles, which can provide better subgoal sequences for the trajectory planner.

First of all, in Reference [5], there are some key definitions.

Heuristic $h(s, s')$ is the octile distance between cells $s$ and $s'$.

**Definition 1.** *An unblocked cell $s$ is a subgoal iff there are two perpendicular cardinal directions $c_1$ and $c_2$ such that $s + c_1 + c_2$ is blocked and $s + c_1$ and $s + c_2$ are not blocked.*

**Definition 2.** *Two cells $s$ and $s'$ are h-reachable iff there is a path of length $h(s, s')$ between them. Two h-reachable cells are safe-h-reachable iff all shortest paths between them are not blocked by obstacles. Two h-reachable cells $s$ and $s'$ are direct-h-reachable iff none of the shortest paths between them contains a subgoal $s'' \notin \{s, s'\}$.*

**Definition 3.** *A subgoal graph $G_S = (V_S, E_S)$ is an undirected graph where $V_S$ is the set of subgoals and $E_S$ is the set of edges connecting direct-h-reachable subgoals. The lengths of edges are the octile distances between subgoals they connect.*

In SSG, there are two steps to find the shortest paths from a start cell to a goal cell. The preprocessing stage abstracts subgoal graphs from the underlying grid maps with alert areas, by identifying subgoals and checking direct-h-reachability. This sparse space representation can eliminate the path symmetry with the strict reachability check and generate subgoals which are not close to each other, as successors. As for the online phase, given the start and goal point, it applies a connect–search–refine approach to return ground-level paths. For our scenarios, the refinement was replaced by LSPI to obtain a smooth and collision-free path. Thus, a connect–search method remained, as shown in Algorithm 1.

Algorithm 1 shows how to find the shortest paths over subgoal graphs. Firstly, the start cell and the goal cell are regarded as subgoals and are connected to subgoal graphs. Then, A* is used to search global abstract paths over modified subgoal graphs. As proven in Reference [5], SSG is a complete and optimal global search algorithm.

In Figure 3d, the grid map with alert areas is 12 × 12. However, the size of $V_S$ of the subgoal graph is 7, and the size of $E_S$ is 7. Obviously, the search space of the subgoal graph is smaller than the grid map's. Moreover, subgoals are placed at the exit of the box canyon on the left side of Figure 3d, and adjacent subgoals are direct-h-reachable, which brings greater security for the trajectory-planning phase.

**Algorithm 1.** Pseudo-code of Simple Subgoal Graphs (SSG).

---

**Searching subgoal graphs**

**function ConnectToGraph**(cell $s$):
if $s \notin V_S$ then
  $V_S = V_S \cup \{s\}$;
  $S \leftarrow$ GetDirectHReachable($s$);
  for all $s' \in S$ do
    $E_S = E_S \cup \{(s, s')\}$;

**function FindAbstractPath**(cells $s, s'$):
ConnectToGraph($s$)
ConnectToGraph($s'$)
$\Pi \leftarrow$ find a shortest path from $s$ to $s'$ over the modified subgoal graph
return $\Pi$;

**function FindPath**(cells $s, s'$)
$\Pi \leftarrow$ TryDirectPath($s, s'$);
if $\Pi \neq$ nopath then
  return $\Pi$;
$\Pi \leftarrow$ FindAbstractPath($s, s'$);
return $\Pi$;

---

*3.4. The LSPI-Based Feasible Trajectory Planner*

3.4.1. LSPI Method

LSPI can deal with continuous state spaces and be more data-efficient and time-efficient than other conventional temporal-different RL methods. Figure 4 shows the overview of the entire LSPI framework [18]. It is a completely off-policy and offline algorithm, and can use sample sets collected arbitrarily from the simulation environment or the real world. At the policy evaluation step, it uses the linear structure to obtain the approximate state-action value function, which is easy to implement and use. At the policy improvement stage, it obtains greedy policies by maximizing state-action values, based on approximate value functions. Furthermore, the transparent inner mechanism is beneficial for users to see how it works and why failures occur.

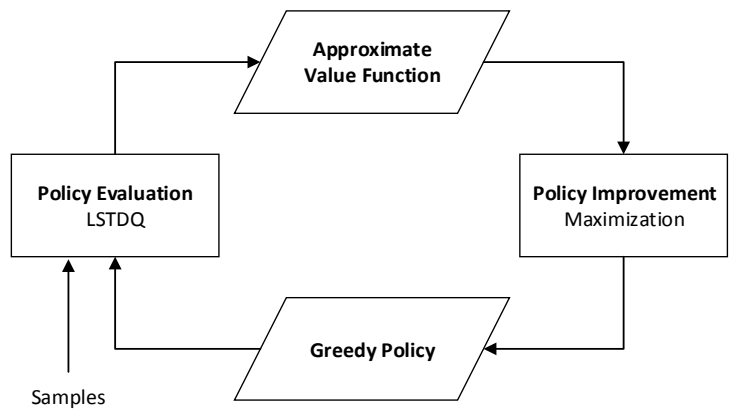

**Figure 4.** The flowchart of least-squares policy iteration. LSTD$Q$: least-Squares temporal-difference learning for the state-action value function

Theoretically, samples from any policy can be collected arbitrarily. Even during policy iteration, samples from the current policy can also be collected and added to train policies. LSPI provides great flexibility for collecting samples, which brings much convenience in practical uses. However, similar to using samples to approximate linear or nonlinear functions, the sample distribution in the state space affects the speed and result of approximating. Therefore, when sample sets are of poor quality, LSPI causes a worse effect via fitting biased distribution based on the data-efficient feature. To solve these problems, in this paper, we not only used random policies to collect samples, but we also tried using random environments to collect samples. Details on how samples are collected are given in Section 4.1.

The prominent advantage of LSPI is its ability to deal with continuous or large state spaces, while tabular RL methods (e.g., Q-learning and Sarsa) cannot solve them, since it is impractical and computationally expensive to use the table structure to store state-action values, (i.e., Computer Go: $10^{170}$ states and Helicopter Control: continuous state spaces). The key of LSPI is that, during the policy evaluation stage, it uses linear architectures to approximate value functions (i.e., uses $\hat{Q}^\pi(s, a; \omega)$ to approximate $Q^\pi(s, a)$ via linear basis functions with free parameters $\omega$). $Q^\pi(s, a)$ is the state-action function under policy $\pi$. The design of basis functions is fundamental work in LSPI. It is popular to use radial basis functions or polynomial basis functions, and we chose polynomial basis functions, since, compared with radial basis functions, polynomial basis functions have clear structures for designers. The definition of $\hat{Q}^\pi(s, a; \omega)$ is

$$\hat{Q}^\pi(s, a; \omega) = \sum_{j=1}^{k} \varphi_j(s, a)\, \omega_j, \tag{2}$$

where $\omega_j$ is the weight parameter, $\varphi_j(s, a)$ is the fixed basis function that is linearly independent, $s$ is the state, $a$ is the action, and $k$ is the number of basis functions.

Two kinds of value-function approximation projections (i.e., Bellman residual minimizing approximation and least-squares fixed-point approximation) are used to approximate value functions. Because learning the Bellman approximation requires "doubled" samples [18] which are impossibly collected from the model-free system, we chose the second approximation that is more practical for the learning task and evaluates the state-action value more accurately.

In the policy-evaluation step, LSTD$Q$ was used to find the solution of least-squares fixed-point approximation which is equivalent to learning $\omega$. Since the model was unknown, samples were used to learn $\omega$ in an incremental way [18].

$$A\omega = b, \tag{3}$$

where matrix $A$ and vector $b$ are

$$A^{t+1} \approx \tilde{A}^{t+1} = \tilde{A}^t + \varphi(s_t, a_t)\big(\varphi(s_t, a_t) - \gamma * \varphi(s'_t, \pi(s'_t))\big)^T, \tag{4}$$

$$b^{t+1} \approx \tilde{b}^{t+1} = \tilde{b}^t + \varphi(s_t, a_t) r_t, \tag{5}$$

where $\gamma$ is the discounted factor, $\pi$ is the policy function, and $r_t$ is the immediate reward.

As seen in the pseudo-code of LSPI [18] in Algorithm 2, during the policy iteration, the same sample sets are used in every iteration, and the iteration ends until $\omega$ is stable. Since sample sets are important for LSPI, the collected sample sets should cover the entire state-action space as uniformly as possible to improve the training effect.

**Algorithm 2.** Pseudo-code of Least-Squares Policy Iteration (LSPI).

| LSPI |
|---|
| **function** LSPI($D, k, \varphi, \gamma, \epsilon, \pi_0$) |
| // $D$: set of samples($s, a, r, s'$) |

```
//  k:      The number of basis functions
//  φ:      Basis functions
//  γ:      Discount factor
//  ϵ:      Stopping criterion
//  ω₀:     initial weight parameters
Ã ← 0
b̃ ← 0
ω' ← ω₀
Repeat
    ω ← ω'
    for each (s, a, r, s') ∈ D
        Ã ← Ã + φ(s, a)( φ(s, a) − γ ∗ φ(s', π(s')))ᵀ
        b̃ ← b̃ + φ(s, a)r
    end
    ω' ← Ã⁻¹b̃
until (|ω − ω'| < ϵ)
return ω
```

3.4.2. Definitions of Subgoal-Approaching (SA) and Obstacle-Avoiding (OA) MDPs

In the feasible-trajectory-planning phase, to increase training efficiency, we decomposed the problem of finding feasible trajectories between subgoals into two sub-problems—approaching subgoals and avoiding obstacles. Since LSPI is used to find a solution to discrete-time MDPs with continuous state spaces, given as a tuple of states, action, rewards, and next states, $(s, a, r, s')$, we separately built two MDPs via abstracting key information from the two sub-problems, to formally describe environments for LSPI. One MDP, named Subgoal-Approaching (SA), describes the process of approaching subgoals with kinematic constraints satisfied. The other MDP, named Obstacle-Avoiding (OA), describes the process of preventing collisions with obstacles. Definitions of SA and OA are described in Table 1. In these two MDPs, we used kinematic equations to update trajectories of the robot, so that the trajectories could preserve G1 continuity. Figure 5 shows the process of planning feasible trajectories. At first, we check the sensor data, and then decide which MDP to execute.

**Table 1.** Definitions of Subgoal-Approaching (SA) and Obstacle-Avoiding (OA) Markov Decision Processes (MDPs).

| Type of MDP | SA MDP | OA MDP |
|---|---|---|
| States | $(d_g, a_g)$ | $(S_1, S_2, S_3, S_4, S_5, S_6)$ |
| Actions | move forward: (0.5, 0.5) <br> turn left: (0.5, 0) <br> turn right: (0, 0.5) | move forward: (0.5, 0.5) <br> turn left: (0.5, 0) <br> turn right: (0, 0.5) |
| Reward | +10 if $d_g < d_n$ <br> $-\tilde{a}_g$ if $\tilde{a}_g > 0$ <br> $1 - \tilde{d}_g - \tilde{a}_g$ else | Comparative value reward in Table 2, where P = 0.9 |

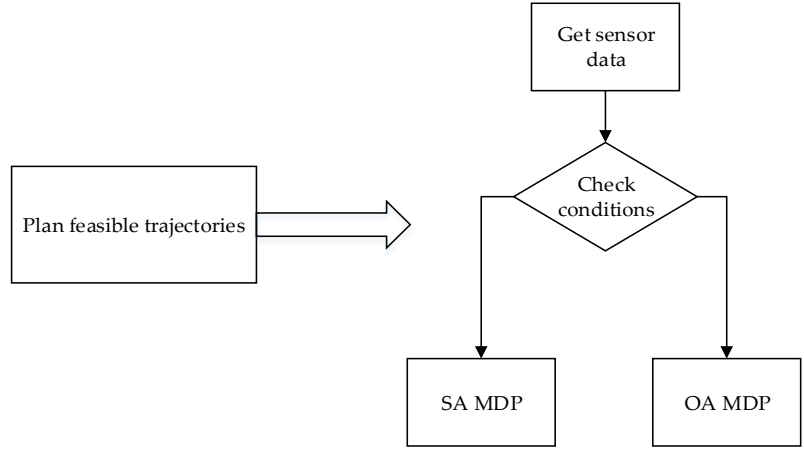

**Figure 5.** The flowchart of the trajectory planner.

In the SA MDP, state $d_g$ denotes the distance between the robot and the goal, state $a_g$ denotes the angle between the robot's orientation and the direction from the robot heading to the goal, and the range of $a_g$ is $[-\pi, \pi]$. When the goal is on the left side of the robot's orientation, the value of $a_g$ is positive. Otherwise, the value is negative. There are three actions (i.e., going forward, turning left, and turning right) that the robot can execute. Each action corresponds to values of $\omega_l$ and $\omega_r$. For example, the action of going forward is equal to setting $\omega_l$ and $\omega_r$ to 0.5. And the action's execution is through the use of kinematic equations to calculate the next positions of the robot via inputting $\omega_l$ and $\omega_r$ (i.e., the angular velocity of left and right tracks). The definition of reward plays an important role in MDP, which is directly related to the training speed and the validity of the resulting policy. When arriving into the predefined acceptable area of subgoals, the robot gets a +10 scalar reward; $d_n$ is the goal tolerance. If $\tilde{a}_g$ is more than 0, the reward function is $-\tilde{a}_g$, where $\tilde{a}_g$ is the absolute value of normalization of $a_g$. Otherwise, the reward function is $1 - \tilde{d}_g - \tilde{a}_g$, where $\tilde{d}_g$ is the normalization of $d_g$.

In the OA MDP, state $S_i$ ($i \in (1,2,3\ldots,6)$) denotes the reading of sensor $i$. The range of $S_i$ is (0, 5). The first three sensors belong to left-side sensors and the others belong to right-side sensors. The setting of action is the same as the SA MDP's. Because of complexities of the obstacle-avoiding process, in general, it is tempting to add prior experience into reward functions, but it may not improve the training effect, which is discussed in Section 3.4.3. To explore the effects of adding prior experience into reward functions, we designed two kinds of reward functions—concise reward and comparative value reward. The first is a simple reward function (i.e., colliding with obstacles means receiving −4, otherwise 0). The comparative value reward adds prior experience about the risk of the current state based on the concise reward, which is shown in Table 2. $S_{min} = min(S_1, S_2, \ldots, S_6)$. $S_{li}$ ($i \in (1,2,3)$) denotes the $i$-th minimum reading of left-side sensors. $S_{ri}$ ($i \in (1,2,3)$) denotes the $i$-th minimum reading of right-side sensors. P is a constant value to tune the reward function. $D_{ur}$ denotes the brake distance for tracked robots. $D_{sa}$ denotes the ideal safe distance for tracked robots keeping away from obstacles. These conditions are checked in order, as shown in Table 2. To increase the performance of learned policies, an action-changing penalty (i.e., when the current action is different from last one, the agent obtains −0.2) can be combined with abovementioned reward functions, which is verified in Section 4.2.1.

**Table 2.** Comparative value reward in obstacle-avoiding MDP.

| Order | Conditions | Annotation | Reward |
|---|---|---|---|
| 1 | $S_{min} < D_{ur}$ | The minimum reading of all sensors is less than the brake distance. | −4 |
| 2 | $S_{min} > D_{sa}$ | The minimum reading of all sensors is greater than the safe distance. | +0 |
| 3 | $S_{min} < D_{sa}$ $S_{l1} \neq S_{r1}$ | The minimum reading of all sensors is less than the safe distance, and the minimum reading of left-side sensors is not equal to the right-side sensor's minimum reading. | $-P * (D_{sa} - min(S_{l1}, S_{r1}))$ |
| 4 | $S_{min} < D_{sa}$ $S_{l2} \neq S_{r2}$ | The minimum reading of all sensors is less than the safe distance, and | $-P * (D_{sa} - min(S_{l2}, S_{r2}))$ |

| | | the second minimum reading of left-side sensors is not equal to the right-side sensor's second minimum reading. | |
|---|---|---|---|
| 5 | $S_{min} < D_{sa}$ | The minimum reading of all sensors is less than the safe distance. | $-P * (D_{sa} - \min(S_{l3}, S_{r3}))$ |

### 3.4.3. Improvements of Training with LSPI

To make the proposed feasible trajectory planner more efficient, we focused on the factors described below, including training environments and reward functions.

Firstly, from the view of input of LSPI, one significant element of successful training is whether sample sets can cover the entire state-action space. It affects the accuracy of the transition probability function, $P: S \times A \times S \to \mathbb{R}$. $P$ denotes given current action $a$, the probability of transferring current state $s$ to next state $s'$.

Sample sets depend on two key factors, which are ways of collecting samples, and training environments. For better sampling, we use not only random policies to ensure that every action is selected with the same probability, but also random positions to make all kinds of different situations happen. When starting the new sampling epoch, the goal position and the robot position are set randomly. We designed two kinds of training environments including simplified office maps and maps with random obstacles. In maps with random obstacles, various collected sample sets contribute to learning the transition probability function, which is proven in Section 4.2.2.

Secondly, scalar reward is important for RL methods. In reinforcement learning, there is a reward hypothesis: realizing goals means the maximization of the expected value of the cumulative sum of a received scalar signal (called reward). The hypothesis is described as follows: [12]

$$G_t = R_{t+1} + \gamma R_{t+2} + \gamma^2 R_{t+3} + \cdots = \sum_{k=0}^{\infty} \gamma^k R_{t+k+1}. \tag{6}$$

It is critical that rewards truly indicate what we want agents to achieve, and generating a reward signal does not depend on knowledge of how the agent chose correct actions. Therefore, the reward signal is not the place to impart to the agent prior knowledge about how to achieve the goal state. When we design the reward function, we should model this function from the brain of intelligent agents, which can realize what designers desire. If environments are well defined, designing the reward function is simple, such as winning +1 and losing 0 in chess. However, if environments are unknown and tasks are complex, it is tempting to add prior experience into reward functions and give the agent a supplemental reward for completing sub-tasks. The reward signal with well-intentioned supplemental rewards may lead agents to behave very differently from what is intended, and agents may end up not achieving the overall goal at all [12]. For example, in OA MDP, the comparative value reward is a reward with well-intentioned supplemental rewards, which is compared with the concise reward in Section 4.2.1, to show the effects of prior experience in reward functions.

## 4. Results

In this section, firstly, we evaluate the performance of SG–RL on static and dynamic maps. Static maps are constructed by laser sensor data from robots that are deployed with the Robot Operating System (ROS). Dynamic maps mean that some uncertainties such as unexpected obstacles appear along the abstract path. Note that unexpected obstacles mean static obstacles which are not known initially because of errors from sensors or environment models. Secondly, we evaluate the effect of reward function and different types of training maps in the obstacle-avoiding MDP. This paper implements the proposed algorithm based on the source code from *http://www2.cs.duke.edu/research/AI/LSPI* and *http://movingai.com/GPPC*. We run experiments on a 3.6-GHz Intel Core i7-6700 Central Processing Unit (CPU) with 16 GB of Random-Access Memory (RAM).

In the simulation experiments, we exploited numerical methods (i.e., solved equations in small steps) to calculate trajectories via kinematic equations. If the step approximates zero, the computed trajectory (a series of discrete trajectory points) will be equivalent to a continuous trajectory computed by analytical methods. However, small steps require more computational resources and also slow down the computation speed. Considering our experiment platform, we chose the step of simulation time as 0.1 s, which is relatively large in order to achieve higher speed and less memory consumption.

*4.1. Performance of the SG–RL Method*

SG–RL is a method used to find rational paths from a start position to a goal position, which decomposes complex path-planning problems into two phases: the first for the global abstract path planning and the second for the feasible trajectory planning.

In the first phase, we used the SSG method to generate subgoal sequences. The time to find subgoal sequences, called H-time, is a key assessment indicator, which can evaluate the ability to eliminate first-move lags. SSG has constrained heading changes, which cannot generate smooth paths for agents moving in continuous environments. By contrast, the proposed hierarchical method, SG–RL, has free heading changes. We evaluated its ability to deal with continuous environments by comparing its path lengths of it with those of SSG.

In the second phase, we used a feasible trajectory planner based on LSPI to find safe paths between subgoals, and all trajectory segments between subgoals could be linked to obtain complete feasible trajectories from the start position to the goal position. In the LSPI method, some parameter values came from Reference [18] and others came from trials. The discounted factor $\gamma$ was 0.9. Basis function $\varphi$ was polynomial. In OA MDP, the order of basis function $\varphi$ was 3, and in SA MDP, the order was 4. The subgoal tolerance was 1.5 and the final goal tolerance was 0.5. The subgoal tolerance was bigger since subgoals only play auxiliary roles in the trajectory planner. The time interval of action choice was 0.5 s. The way of collecting samples was to choose actions randomly per second. If robots collided with obstacles or boundaries, this sampling epoch ended and robots started from random positions.

Action-switching frequency is a significant factor which can evaluate the performance of planned trajectories. Its definition is the ratio of the sum of the current action that is different from the last action to all actions. For robots, it cannot be high because of limitations of hardware systems and network latency. Realizing fast and frequent responses requires more precise chassis gears and more energy. Meanwhile, in areas with signal interference, network latency is relatively high.

4.1.1. Performance on Static Maps

We evaluated the SG–RL method with four maps described in Figure 6. We trained the robot to learn to avoid obstacles with LSPI in Figure 6a. The three maps for navigation tasks (Figures 6b–d) were between 250 and 521 times larger than the training map. Since the resolution of these three original maps (i.e., 2.5 cm per pixel) was too high for doing experiments on them, a down-sampling method was used to compress them. These original maps were composed of grayscale information, and we transformed them into grid maps (if there existed an obstacle in a cell, then the value of this cell was set to 1, otherwise 0).

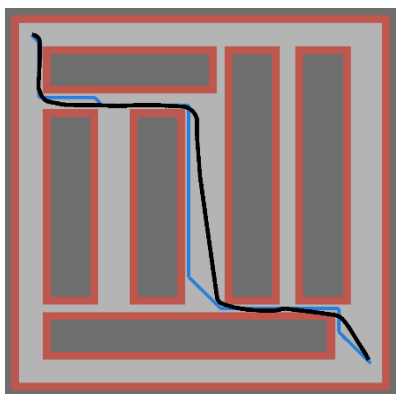
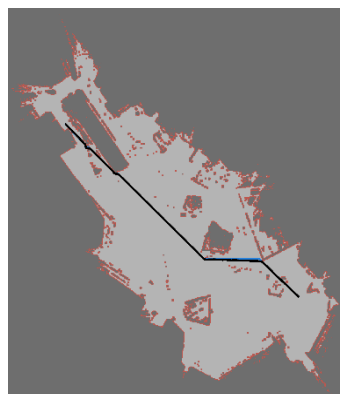

(**a**) Training environment—50 by 50  (**b**) Building 1—820 by 820

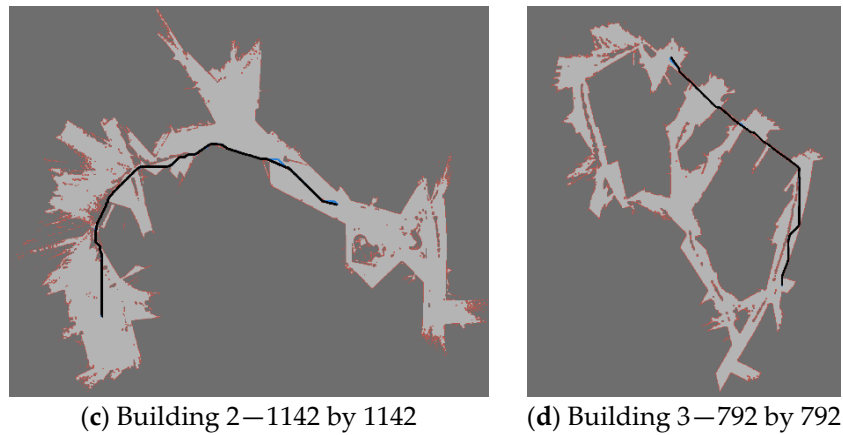

(**c**) Building 2—1142 by 1142  (**d**) Building 3—792 by 792

**Figure 6.** (**a**) The training environment was a simplified office. (**b**–**d**) Given the start position and the goal position, the blue line is the path planned by the Simple Subgoal Graphs (SSG), and the black line is the trajectory planned by SG–RL. The dark-gray region is unpassable while the light-gray region is traversable for robots. Red regions deemed close to obstacles denote alert areas.

Table 3 shows the action-switching frequency of the robot completing the whole path on different maps. Since the learned policy was deterministic, this indicator was also fixed given the same start–goal location pair. All these values were below 10%, which indicates that the robot maneuvered by SG-RL can work well in large and realistic environments with low action-switching frequencies. In Table 4 the path length of SG–RL in Building 1 was almost equal to that in Building 3; however, in Table 3, the action-switching frequency of Building 3 was larger than that of Building 1, since there were more narrow passages in Building 3, where the robot needed to quickly change its actions for adapting to the complex environment (i.e., the robot changed its action to traverse many left and right turns). As a consequence, as the complexity of test maps increased, the action changes that were essential for the environment accordingly increased in number, which finally, increased the action-switching frequency. Action-switching frequency is also related to the performance of learned policy. If the performance of learned policies improves, the evaluation factor will decrease. Therefore, in Section 4.2, we present two ways to improve the performance of learned policies (i.e., by modifying reward functions and by choosing suitable training maps).

**Table 3.** Action-switching frequency of SG-RL.

| Method | Building 1 | Building 2 | Building 3 |
| --- | --- | --- | --- |
| SG-RL | 0.0492 | 0.0849 | 0.0827 |

As shown in Table 4, although SSG could obtain an optimal path over the grid map which was the discretization of continuous environments, SG–RL could plan even shorter paths, showing that SG–RL can deal with continuous environments and plan trajectories following kinematical equations. Briefly, SG–RL without constrained heading changes can work well in continuous environments, which is suitable for robots navigating in the real world.

**Table 4.** Path length of SSG and SG–RL.

| Method | Building 1 | Building 2 | Building 3 |
| --- | --- | --- | --- |
| SSG | 563.23 | 958.72 | 587.58 |
| SG–RL | 558.18 | 931.38 | 574.68 |

Table 5 shows, given the start position and the goal, the average value of H-time of A* and SG–RL on different maps. The H-time of A* is 195 times that of SG–RL on the Building 2 map and 107 times that of SG–RL on the Building 3 map. The H-time of SG–RL was below 1 millisecond on large-scale maps, which can meet the practical requirements and eliminate first-move lag. SG–RL used SSG with great success to obtain subgoal sequences, which benefitted from reducing the search space by abstracting topological structures of maps.

Table 5. H-time of A* and SG–RL.

| Method | Building 1 (ms) | Building 2 (ms) | Building 3 (ms) |
|---|---|---|---|
| SG–RL | 0.117 | 0.541 | 0.371 |
| A* | 20.828 | 105.582 | 39.860 |

4.1.2. Performance on Dynamic Maps

To evaluate SG–RL's ability to deal with uncertainties in environments, we added some random shapes of obstacles along the subgoal sequence. Certainly, by reconstructing the subgoal graphs and replanning abstract paths, SG–RL could still cope with the new environment. However, the preprocessing stage plus the path replanning needs some time, which is not suitable for use online. To tackle the problem elegantly, SG–RL continues using the initial subgoal sequence, but deals locally with unexpected situations. Owing to the capacity of generalizability, SG–RL behaves well enough in uncertain environments, as demonstrated in Figure 7. Even though there were some obstacles blocked in the path between subgoals, the robot controlled by SG–RL could achieve non-collision moves.

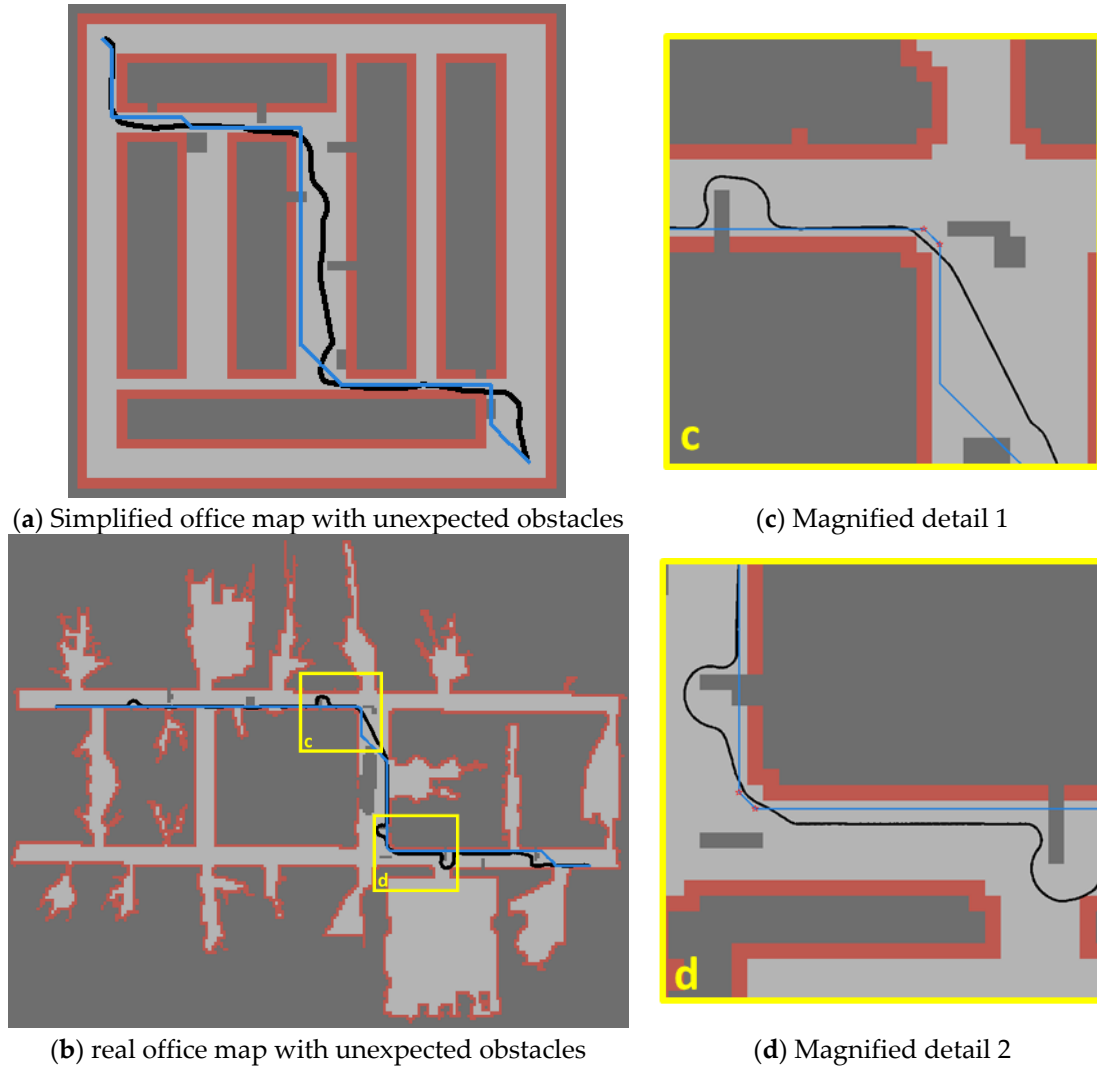

(**a**) Simplified office map with unexpected obstacles  (**c**) Magnified detail 1
(**b**) real office map with unexpected obstacles  (**d**) Magnified detail 2

**Figure 7.** (**a**,**b**) The blue line is the path planned by SSG based on the map without unexpected obstacles. The black line is the trajectory planned by SG–RL without replanning the subgoal sequence. The dark-gray region is unpassable, while the light-gray region is traversable for robots. Red regions deemed close to obstacles denote alert areas. (**c**,**d**) Two magnified details.

## 4.2. Influence Factors of the Obstacle-Avoiding MDP

During the second phase, SG–RL formalizes the problem of finding smooth and collision-free trajectories into two MDPs. OA MDP is an important part of the feasible trajectory planner, and it is difficult to train this MDP since it needs to deal with complex states. Therefore, it is essential to research its influence factors like reward functions and training environments.

### 4.2.1. Reward Functions

Firstly, to explore the effects of prior experience in reward functions, we compared the concise reward with the comparative value reward. The comparative value reward consists of the concise reward and some supplemental rewards concerning the risk of the current state. Secondly, to reduce action-switching frequencies, we added action-changing punishment into the original reward functions.

To reduce the randomness of training, we trained 100 times and randomly collected 60,000 samples each time. To evaluate the learned obstacle-avoiding policy, we let the robot move from the start position to the finish line on the test map (Figure 8). If the robot collided with obstacles, the policy was considered as a failure, otherwise it was considered a success.

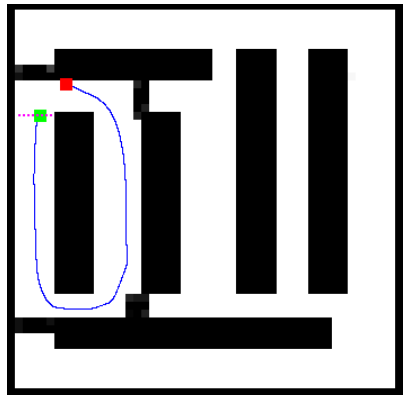

**Figure 8.** This is a successful sample of one robot with learned obstacle-avoiding policy completing the test. The red square is the start position, and the pink dotted line is the finish line. The green square is the position that the robot arrived at the finish line.

As seen in Table 6, the ratio of successful OA policies with the concise reward is five times higher than that with the comparative value reward. The other evaluation indicator of concise reward is over five times higher than that of the comparative value reward. Obviously, training with the concise reward can more easily get an obstacle avoidance policy than training with the comparative value reward. The comparative value reward including prior experience makes rewards dense; however, meanwhile, it brings difficulties for training. Therefore, training with the concise reward can perform better, and adding prior experience into reward functions may bring some negative effects on training. The better place to impart to the agent prior experience may be the initialization.

**Table 6.** Comparison of concise reward and comparative value reward.

| Reward Function | Ratio of Successful OA Policies | Ratio of Policies with Action-Switching Frequency Below 30% |
| --- | --- | --- |
| concise reward | 45% | 38% |
| comparative value reward | 9% | 7% |

As seen in Table 6, the learned policies did not perform well in the aspect of action-switching frequencies; thus, we brought the punishment concerning action changes into reward functions to reduce action-switching frequencies while avoiding obstacles.

The punishment concerning action changes suggests that when current action is different from last one, the agent obtains a value of −0.2. Then, the entire combined reward has "strong" requirements (i.e., avoid obstacles) and "weak" requirements (i.e., reduce action changes) which means that, after "strong" requirements are satisfied, "weak" requirements can be satisfied as much as possible.

Shown in Table 7, the evaluation factor is the ratio of policies with action-switching frequency below 30%. Obviously, original reward functions with action-changing punishment can increase this evaluation factor. Therefore, action-changing punishment does a great favor to both the concise reward and the comparative value reward, reducing the action-switching frequency by increasing the performance of learned policies, thereby also demonstrating that the combined rewards work better than separate ones.

**Table 7.** Effect of adding action-changing punishment.

| Reward Function | Without Action-Changing Punishment | With Action-Changing Punishment |
|---|---|---|
| Concise reward | 38% | 49% |
| Comparative value reward | 7% | 14% |

4.2.2. Different Types of Training Maps

In this section, there are two types of training maps. The first type is the simplified office map (Figure 7a). The second type is the map with random obstacles, whose size is the same as the simplified office map. To build the second-type map, we firstly set 0.05 as the ratio of obstacles in the whole map. We randomly set cells as obstacles one by one under the same probability until the ratio of obstacles reached 0.05. Random obstacles were uniformly distributed in the space represented in the map. The test map was the same as the first type. The reward function we used was a concise reward with action-changing punishment.

**Table 8.** Comparison of different types of training maps.

| Reward Function | Ratio of Successful OA Policies | Ratio of Policies with Action-Switching Frequency Below 30% | Average Iteration Numbers of LSPI |
|---|---|---|---|
| Simplified office map | 53% | 48% | 7.94 |
| Map with random obstacles | 89% | 57% | 6.01 |

As shown in Table 8, using the second type of map can more easily obtain obstacle-avoiding policies than using the first type. On the map with random obstacles, the robot chose action randomly, which realized not only randomness in action selection, but also in training environments. It can be seen that sample sets collected from the second-type map could cover more state-action spaces than sample sets from the first type, which could help LSPI to converge useful policies. Applying the second-type map could decrease the action-switching frequency as a whole. Using the second type meant fewer iteration numbers, which could speed up the training process. The experimental result shows that using the second type of environment could not only collect more different sample sets, but could also increase the accuracy of the learned transition probability function, so as to increase the performance of learned policies and decrease the action-switching frequency.

**5. Conclusions**

In this paper, we proposed a hierarchical path-planning framework called SG–RL in continuous and uncertain environments. By "rational", we mean (1) efficient path planning to eliminate first-move lag; (2) collision-free and smooth for agents with kinematic constraints satisfied. SG–RL plans rational paths in a two-phase manner (i.e., the first phase for global abstract path planning and the second for feasible trajectory planning). In the first phase, SG–RL uses SSG to generate subgoal sequences efficiently for eliminating first-move lag, and overcome sparse reward and local minima trap for LSPI. In the second phase, following kinematical equations, SG–RL uses LSPI to plan feasible trajectories for agents between adjacent subgoals, and handles unexpected obstacles without replanning subgoal sequences.

In simulation experiments, firstly, SG–RL was evaluated on realistic and large-scale maps constructed by laser sensor data. The action-switching frequency was below 10% and the time to obtain subgoal sequences was between 195 times and 107 times faster than A* on grid maps. The path length of SG–RL was shorter than that of SSG. Secondly, evaluated on dynamic maps, SG–RL can deal with uncertainties based on original subgoal sequences without reconstructing subgoal graphs from the changed map.

We demonstrated that the design of reward functions and the type of training environment are important factors for learning feasible policies. A concise reward with action-changing punishment achieved the best results in the obstacle-avoiding MDP, which means that reward functions with prior experience may bring difficulties for agents to learn effective policies. The map with random obstacles increased the accuracy of the learned transition probability function, whose ratio of getting an effective obstacle avoidance policy was 89%. SG–RL, as a hierarchical path-planning method, combines the strengths of SSG and LSPI to overcome their limitations, greatly improving the efficiency of finding rational paths.

In future work, continuous action spaces can be taken into consideration. Specifically, the range of angular velocity can be [−0.5, 0.5], which means that robots can do more free and complex actions, such as going backward. In the second phase of SG–RL, if the number of state-action pairs is not huge, LSPI can keep the relative size of Q-value for every state-action pair to ensure the right action choice. However, since adding continuous action space will greatly increase the state-action space, it is difficult for linear basis functions to evaluate the Q-value precisely. Inspired by the idea of asynchronous advantage Actor-Critic [30], we can use an artificial neural network to approximate state-action value functions, and use another network to approximate policy functions, which can increase the accuracy of evaluating the Q-value.

**Author Contributions:** J.Z. and L.Q. conceived and designed the paper structure and the experiments; J.Z. performed the experiments; Y.H., Q.Y., and C.H. contributed with materials and analysis tools.

**Funding:** This work was sponsored by the National Science Foundation (NSF) project 61473300, China.

**Acknowledgments:** The work described in this paper was sponsored by the National Natural Science Foundation of China under Grant No. 61473300. We appreciate the fruitful discussion with the Sim812 group: Qi Zhang and Yabing Zha.

**Conflicts of Interest**: The authors declare no conflicts of interest. The founding sponsors had no role in the design of the study; in the collection, analysis, or interpretation of data; in the writing of the manuscript, and in the decision to publish the results.